\documentclass[final,5p,times]{elsarticle}
\usepackage{amssymb}
\usepackage[nodots]{numcompress}

\usepackage{lineno}

\usepackage{multirow}

\journal{Computers \& Graphics}

\begin{document}

\begin{frontmatter}



\title{Facial Expressions Tracking and Recognition: Database Protocols for Systems Validation and Evaluation}


\begin{abstract}
	
Each human face is unique. It has its own shape, topology, and distinguishing features. As such, developing and testing facial tracking systems are challenging tasks. The existing face recognition and tracking algorithms in Computer Vision mainly specify concrete situations according to particular goals and applications, requiring validation methodologies with data that fits their purposes. However, a database that covers all possible variations of external and factors does not exist, increasing researchers' work in acquiring their own data or compiling groups of databases.

To address this shortcoming, we propose a methodology for facial data acquisition through definition of fundamental variables, such as subject characteristics, acquisition hardware, and performance parameters. Following this methodology, we also propose two protocols that allow the capturing of facial behaviors under uncontrolled and real-life situations. As validation, we executed both protocols which lead to creation of two sample databases: FdMiee (Facial database with Multi input, expressions, and environments) and FACIA (Facial Multimodal database driven by emotional induced acting).

Using different types of hardware, FdMiee captures facial information under environmental and  facial behaviors variations. FACIA is an extension of FdMiee introducing a pipeline to acquire additional facial behaviors and speech using an emotion-acting method. Therefore, this work eases the creation of adaptable database according to algorithm's requirements and applications, leading to simplified validation and testing processes.

\end{abstract}


\end{frontmatter}

%\linenumbers

\section{Introduction}

In the field of Computer Vision (CV), there are several existing databases that contain a wide range of facial expressions and behaviors, developed based on specific scenarios. The data contained in these databases is usually used for validation and performance tests, as well as training of facial models in CV algorithms \cite{yin2008high,pantic2005web,tolba2006face}. To date, computational works include only a limited number of features though, representing typical facial extraction elements \cite{Kapoor2003,Cheon2009,Fischer2004}. In fact, there is no single database that integrates a full set of situations: some are dedicated only to expressions, others to lighting conditions, some are just for extracting facial patterns used to define training models, others for emotion classification, etc. This means the information is split across a variety of databases, making it impossible to validate a facial tracking system under numerous specific situations (e.g. partial face occlusions from hardware or glasses, changes in background, variations in illumination, head pose variations, etc...) or train emotion classifier systems capable of capturing the subtleties of the face using only one database. This drawback usually leads to systems over-fitting to data, presenting a high specificity to a certain environment or limiting facial features recognized \cite{baggio2012mastering}. Therefore, every time it is required to design validation and performance tests or training sets, researchers struggle to find databases that fit all system's requirements \cite{baggio2012mastering}. As example, to deploy the recent face tracking system \cite{Cao2014} it was needed the compilation of three different databases. In alternative, researchers define and setup their own procedures to acquire own databases, collecting subjects, defining protocols, and preparing  capture equipment - which are all time-consuming processes. This "database customization" requirement exists since databases require specific features or formats (e.g. high-resolution videos and infra-red pictures) according to CV system's profile and goal. These features and formats contain a wide range of variations in external and facial behavior parameters to simulate real-life situations and provide information that would reproduce the scenario accurately where the system is going to be applied \cite{baggio2012mastering}.

In this work, we designed two generic protocols and developed a methodology for data acquisition for face recognition systems, as well as for tracking and training of CV algorithms. Our methodology defines each acquisition protocol to be composed of three basic variables: \textit{i)} subject characteristics, \textit{ii)} acquisition hardware and \textit{iii)} performance parameters. These variables are classified as flexible (i.e. can be altered according to system requirements, not influencing protocol guidelines) and fixed (i.e. defined and constrained b the protocol guidelines). The flexible variables are connected to system requirements, and the fixed ones to the information recorded and simulated. As performance variables, we define the following parameters: external (e.g. environment changes in lightning and background) and facial (e.g. variations in facial expressions and their intensity). To test the accuracy and performance of algorithms in facial features tracking or to train face models, used databases need to contain a broad set of external and facial behavior variations. Setting up these variables through our proposed methodology and adopting our protocols eases the process of acquisition of databases with facial information under real-life scenarios and realistic facial behaviors. To validate this process, we followed both protocols and acquired two sample databases. We also analysed the obtained results to establish proof-of-concept.

We dubbed the first protocol \textbf{Protocol I} that generated \textit{FdMiee} ``Facial database with multi input, expressions and environments''. Protocol I aims to guide researchers through acquiring data using three capture hardware while varying the performance variable, giving special focus to external parameters variation. As Protocol I's extension, \textbf{Protocol II} introduces variations in performance variable regarding facial behaviors. Validation of this protocol generated \textit{FACIA} ``Facial Multimodal database driven by emotional induced acting''.

\begin{figure}
	\centerline{\includegraphics[width=\linewidth]{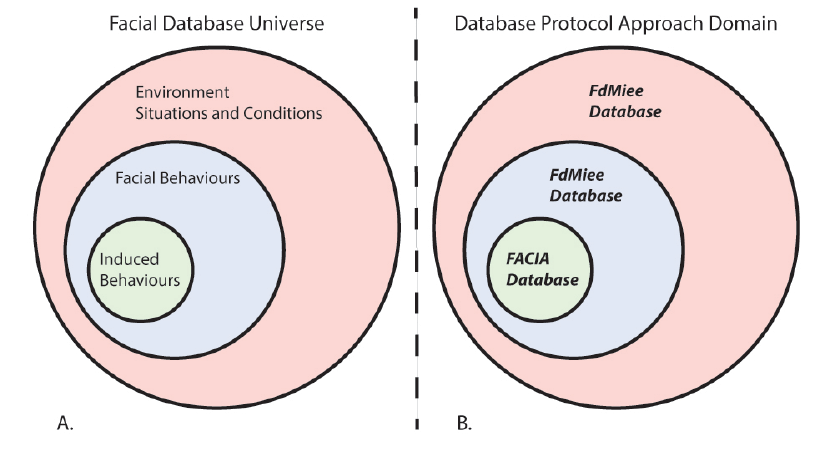}}
	\caption{Summary of database protocols' contributions (B) in facial database universe (A).}
	\label{imgFacialDatabasesWorld}
\end{figure}

Figure \ref{imgFacialDatabasesWorld} represents our overall contribution schematically, regarding the types of data captured in the protocols. It represents the Facial databases' universe through Environment situations and conditions, where we include the group of available facial behaviors, with a small part reserved to introduce behaviors (Figure \ref{imgFacialDatabasesWorld} - A). Taking this scheme into account, we can mirror the domain of our database protocol and represent the contributions of FdMiee and FACIA diagrammatically (Figure \ref{imgFacialDatabasesWorld} - B). 

\section{Background}

To develop guidelines for database acquisition, we researched the literature for methodologies and variance parameters required to test and evaluate CV systems. We analyzed state-of-the-art databases, and classified them into two groups, according to their output format: video and image-based. The most commonly-used video databases are as follows:

\begin{itemize}
	\item BU-4DFE (3D capture + temporal information): A 3D Dynamic Facial Expression Database \cite{yin2008high};
	\item BP4D-Spontaneous: a high-resolution spontaneous 3D dynamic facial expression database \cite{Zhang2014692};
	\item MMI Facial Expression Database \cite{pantic2005web};
	\item VidTIMIT Audio-Video Database \cite{sanderson2009multi};
	\item Face Video Database of the Max Plank Institute \cite{CyberData}.
\end{itemize}

Comprehensive and well-documented video databases exist, for example \cite{CMUFIA} and \cite{messer1999xm2vtsdb}. However, to access them, a very strict license must be procured and a payment provided. BU-4DFE \cite{yin2008high} presents a high-resolution 3D dynamic facial expression  database. Facial expressions are captured at 25 frames-per-second while performing six basic Ekman's emotions. Each expression sequence contains about 100 frames spread through 101 subjects. More recently, this database was extended to create a 3D spontaneous facial expressions \cite{Zhang2014692}. Another facial expressions database commonly used is the MMI database \cite{pantic2005web}. It is an ongoing project that holds over 2000 videos and more than 500 images from 50 subjects. Also information of displayed AU's is given with the samples. The VidTIMIT Audio-Video  \cite{sanderson2009multi} contains video and audio recordings from 43 people reciting 10 short sentences per person. Each person also performs a head rotation sequence per session, which in facial recognition can allow pose independence. Finally, Face Video Database from the Max Planck Institute provides videos of facial action units, used for Face and Object Recognition, though no more information is given \cite{CyberData}. Usage of videos instead of images on the model training allows a better detection of spontaneous and subtle facial movements. However, available databases are limited to standard facial expressions detection \cite{yin2008high,pantic2005web} or do not explore situations with different lighting levels. 

Regarding image-based databases, we came across a comparison study in the table VIII of \cite{tolba2006face}. This table describes the commonly-used image-based databases for validation of face tracking systems. It also exposes their limitations. As examples of current image-based databases, we analyzed the following databases:

\begin{itemize}
	\item Yale \cite{belhumeur1997eigenfaces};
	\item Yale B \cite{KCLee05};
	\item the FERET \cite{phillips2000feret};
	\item CMU Pose, Illumination and Expression (PIE) \cite{sim2002cmu};
	\item Oulu Physics \cite{marszalec2000physics}.
\end{itemize}

Regarding \emph{Yale} \cite{belhumeur1997eigenfaces} and \emph{Yale B} \cite{KCLee05} database, it contains a limited number of grayscale images with well-documented variations on lighting, facial expressions, and pose variations. In contrast, the \emph{ FERET} database  \cite{phillips2000feret} has a high number of subjects with a complete pose variation. However, no information about lighting is given. Another interesting database is the \emph{CMU PIE} \cite{sim2002cmu} which also tests extreme lighting variations for 68 subjects. These three databases are frequently used for facial recognition, not only for model training but also for validation. Finally, we also highlight the \emph{Oulu 
	Physics} \cite{marszalec2000physics} database, since it presents a variation on lighting color (horizon, incandescent, fluorescent, and daylight) on 125 faces.

Based on this research, we concluded that there is a wide range of databases that explore and simulate diverse facial expressions under different environment conditions. However, the available information is spread throughout many databases. In other words, a single database that combines all these facial and environment behaviors and variations providing a complete tool for validation of facial expressions tracking and classification is still non-existent.

In \cite {douglas2007humaine}, a complete state-of-the-art on emotional databases available nowadays can be found. We searched for a facial expressions database that would simultaneously provide color and depth video (3D data stream) as well as speech information, along with emotional data. Our search criteria, however, were not fulfilled.

The increase of affect recognition CV methods \cite{zeng2009survey} lead to a necessity of databases generation containing spontaneous expressions. To establish how to induce these expressions in participants, we analyzed the review paper on Mood Induction Procedures (MIP's) \cite{gerrards1994experimental} and investigated which resource materials could be used to enhance and introduce realism in expressed emotions \cite{martin2006enterface}. We concluded that the most commonly-used emotion induction procedure is the Velten method, characterized by a self-referent statement technique. However, the most powerful techniques are combinations of different MIPs, such as Imagination, Movies/Films instructions or Music \cite{gerrards1994experimental} . Therefore, the technique chosen for our experiment was a combination of the Velten technique with imagination, where we proposed an emotional sentence enacting, similar to the one presented by Martin \emph{et al.} \cite{martin2006enterface}.

Some available databases that use similar MIP's induce emotions in the users by asking them to imagine themselves in certain and pre-defined situations \cite{wilting2006real,velten1968laboratory}. However, the usage of this procedure without complementary material (e.g. sentences) does not guarantee facial expressivity from the user \cite{wilting2006real,gerrards1994experimental}. Since we intended to record speech, we analysed state-of-the-art multimodal databases \cite{douglas2007humaine} and found that there was none containing Portuguese speech. Therefore, we decided to explore this potential research avenue.

\section{Protocol Methodology}

Analysing the background and details of facial data acquisition setups, we propose that to create a protocol, three fundamental variables need to be characterized: \textit{subject characteristics}, \textit{acquisition hardware} and \textit{performance parameters} (Table \ref{tblDefinitionProtocol}).  

%\begin{figure}[ht]
% \centering
% \includegraphics[width=3.5in]{DefinitionProtocol}
% \caption{Protocol flexible and fixed variables.}
%  \label{imgDefinitionProtocol}
%\end{figure}

\begin{table}%
	\caption{Protocol flexible and fixed variables.\label{tblDefinitionProtocol}}{%
		\begin{tabular}{lll}
			\hline
			\multicolumn{3}{|c|}{\textbf{Protocol Variables}}                                                                          \\ \hline \hline
			\multicolumn{2}{|c|}{\textbf{Flexible}}                                        & \multicolumn{1}{c|}{\textbf{Fixed}}       \\ \hline
			\multicolumn{1}{|c|}{Subjects}        & \multicolumn{1}{c|}{Acquisition}       & \multicolumn{1}{c|}{Performance}          \\
			\multicolumn{1}{|c|}{Characteristics} & \multicolumn{1}{c|}{Hardware}          & \multicolumn{1}{c|}{Parameters}           \\ \hline
			\multicolumn{1}{|l|}{Gender}          & \multicolumn{1}{l|}{Webcam}            & \multicolumn{1}{l|}{External Parameters:} \\
			\multicolumn{1}{|l|}{Age}             & \multicolumn{1}{l|}{HD Camera}         & \multicolumn{1}{l|}{\hspace{5mm}Background}           \\
			\multicolumn{1}{|l|}{Race}            & \multicolumn{1}{l|}{Infra-Red Camera}  & \multicolumn{1}{l|}{\hspace{5mm}Lightning}            \\
			\multicolumn{1}{|l|}{(...)}           & \multicolumn{1}{l|}{Microsoft Kinect}  & \multicolumn{1}{l|}{\hspace{5mm}Multi-Subject}        \\
			\multicolumn{1}{|l|}{}                & \multicolumn{1}{l|}{High-Speed Camera} & \multicolumn{1}{l|}{\hspace{5mm}Occlusions}           \\
			\multicolumn{1}{|l|}{}                & \multicolumn{1}{l|}{(...)}             & \multicolumn{1}{l|}{Facial Parameters:}   \\
			\multicolumn{1}{|l|}{}                & \multicolumn{1}{l|}{}                  & \multicolumn{1}{l|}{\hspace{5mm}Head Rotation}        \\
			\multicolumn{1}{|l|}{}                & \multicolumn{1}{l|}{}                  & \multicolumn{1}{l|}{\hspace{5mm}Expressions:}         \\
			\multicolumn{1}{|l|}{}                & \multicolumn{1}{l|}{}                  & \multicolumn{1}{l|}{\hspace{10mm}Macro}                \\
			\multicolumn{1}{|l|}{}                & \multicolumn{1}{l|}{}                  & \multicolumn{1}{l|}{\hspace{10mm}Micro}                \\
			\multicolumn{1}{|l|}{}                & \multicolumn{1}{l|}{}                  & \multicolumn{1}{l|}{\hspace{10mm}False}                \\
			\multicolumn{1}{|l|}{}                & \multicolumn{1}{l|}{}                  & \multicolumn{1}{l|}{\hspace{10mm}Masked}               \\
			\multicolumn{1}{|l|}{}                & \multicolumn{1}{l|}{}                  & \multicolumn{1}{l|}{\hspace{10mm}Subtle}               \\
			\multicolumn{1}{|l|}{}                & \multicolumn{1}{l|}{}                  & \multicolumn{1}{l|}{\hspace{10mm}Speech}               \\ \hline
		\end{tabular}
	}
	%		\begin{tabnote}%
	%			\Note{Source:}{This is a table
	%				sourcenote. This is a table sourcenote. This is a table
	%				sourcenote.}
	%			\vskip2pt
	%			\Note{Note:}{This is a table footnote.}
	%			\tabnoteentry{$^a$}{This is a table footnote. This is a
	%				table footnote. This is a table footnote.}
	%		\end{tabnote}%
\end{table}%

These variables are classified as being either flexible or fixed, according to their impact on the protocol guidelines. Subject characteristics and acquisition hardware are flexible variables, as they can be changed according to system requirements. For example, use male subjects captured with a high-speed camera or other kind of hardware available, since they do not influence the guidelines of acquisition itself, but only interfere with the acquisition setup. In contrast, fixed variables such as performance parameters, influence guidelines definitions, i.e. different performance parameters require us to take different steps for their simulation and acquisition.

\textit{Subject characteristics} include gender, age, race, and other features that can be extrapolated from the subjects' samples. This variable introduces specific facial behaviors (e.g. cultural variations in emotion expressions) in the database. Regarding, \textit{acquisition hardware}, we enabled the usage of any type of input hardware according to acquisition specifications. Different combinations of these flexible variables can be applied to any of the fixed \textit{performance parameters} guidelines. Performance variables describe the procedures for acquiring the data required for performance tests of CV algorithms. They are split into External and Facial categories, according to what we want to test. External parameters are related to changes in the environment, such as background, lightning, number of persons in a scene (i.e. multi-subject), and occlusions \cite{Cotter2010,Buciu2005,Bourel2001}. These variables are almost infinite \cite{huang2007labeled} due to their uncontrolled nature in real-life environments. Facial behaviours should contain facial expressions data triggered by emotions, such as macro, micro, subtle, false, and masked expressions \cite{ekman,martin2009philosophy,buxton2013impaired} or even speech information. Ekman \textit{et al.} \cite{ekman} defines six universal emotions: anger, fear, sadness, disgust, surprise and happiness. These universal emotions are expressed in different ways according to a person's mood and intentions. The way they are expressed leads us to an expressions-classification:

\begin{itemize}
	\item \textbf{Macro}: These expressions last between half a second and 4 seconds. They often repeat and fit what is being said as well as the speech. Facial expressions of high intensity are usually connected to six universal emotions \cite{ekman,martin2009philosophy};
	\item \textbf{Micro}: Brief facial expressions (e.g. milliseconds) related to emotion suppression or repression \cite{ekman,martin2009philosophy};
	\item \textbf{False}:  Mirrors an emotion that is deliberately performed, ans is not being felt \cite{ekman,martin2009philosophy};
	\item \textbf{Masked}: False expression created to mask a felt macro-expression \cite{ekman,martin2009philosophy};
	\item \textbf{Subtle}: Expressions of low intensity that occur when a person starts to feel an emotion or shows an emotional response to a certain situation, another person, or surrounding environment. This is usually of low intensity \cite{buxton2013impaired}. 
\end{itemize}

Facial behaviors generated by speech usually contain a combination of the above expressions \cite{Koolagudi2012sim}.

Following this methodology, we developed two protocols. We dubbed the first protocol to generate \textit{FdMiee} \textbf{Protocol I}. To validate this protocol, we acquired data from eight subjects with different characteristics. We applied low-resolution, high-resolution, and Infra-red cameras as acquisition hardware variables. As performance parameter variables, we simulated multi-input expressions and environments to test the invariance and accuracy of facial tracking systems exposed to changes, e.g. different lighting conditions, universal-based and speech facial expressions. To validate the results, we executed 360 acquisitions and demonstrated the protocol's potentials to acquire data containing uncontrolled scenarios and facial behaviors. We dubbed the second protocol to create \textit{FACIA} database \textbf{Protocol II}. This is an extension of Protocol I's performance parameters variables, introducing induced facial behaviors. To validate the results, we studied the protocol's effectiveness for acquiring multimodal databases of induced facial expressions with speech, color, and depth video (3D data stream) data. To achieve this validation goal, we presented a novel induction method using emotional acting to generate facial behaviors inherent to expressions. We also provided emotional speech in the Portuguese language, since currently there is not any 3D facial database that uses this language. Similar to \textit{FdMiee}, in \textit{FACIA} we created proof-of-concept through an experiment with eighteen participants, in a total of 504 acquisitions.

As a typical protocols' usage example, a research team has available database of 10 female subjects aged between 20-22. They would like to compile a database to test the head rotation tracking accuracy of a CV algorithm using a HD camera. Therefore, they define as \textit{subject characteristics} the female gender and age range. Then, they choose a HD camera as \textit{acquisition hardware} and afterward need to pick the Facial parameter: head rotation as Performance parameter. Finally, they need to follow our validated FdMiee protocol.

In summary, to follow the protocols, we first choose the parameters to simulate as fixed Performance variables. This allow us to define the acquisition guidelines. Secondly, we determine the hardware variable and generate an acquisition setup. It is important to note that this variable is flexible, and thus changing this variable will not impact the guidelines. The same is verified using different subject characteristics.

\section{Protocols and Validation}

In this section, we describe in detail the two protocols that follow our proposed methodology. \textbf{Protocol I} resulted in the FdMiee sample database that contains facial data from uncontrolled scenarios. FdMiee focuses essentially on performance variable guidelines of external parameters. The obtained data was recorded with three types of acquisition hardware. As its extension, \textbf{Protocol II} focuses on testing and simulation of facial parameters of the performance variables, using Microsoft Kinect as hardware.

\subsection{Protocol I}

Facial recognition and tracking systems are highly dependent on external conditions (i.e. environment changes) \cite{baggio2012mastering}. To reduce this dependency, we developed a protocol based on our proposed methodology, for database creation with changes in terms of external parameters, such as light, background, occlusions, and multi-subject. For facial parameters, we setup guidelines to capture variations in head rotation, as well as universal-based, contempt and speech facial expressions. Table \ref{tblDefinitionProtocol} summarizes the performance parameters acquired through this protocol.

\subsubsection{Requirements}

As protocol requirements, we setup the acquisition hardware and equipment to simulate the selected external and facial parameters.

\subsubsection*{Acquisition Hardware}

The chosen acquisition hardware simulates realistic scenarios captured using three types of hardware. To test the protocol guidelines, we chose the following equipment:

\begin{itemize}
	\item Low-Resolution (LR) camera
	\item High-Resolution (HR) camera
	\item Infra-Red (IR) camera
\end{itemize}

The first two cameras (LR and HR) allow us to study the influence of resolution on face tracking, face recognition, and expression recognition \cite{tian2004evaluation}. The IR camera allows us to disregard lighting variations \cite{wolff2003quantitative,jafri2009survey,singh2004infrared} and provides a different kind of information than HR and LR cameras. The hardware used in this protocol should be aligned with one another to ensure future comparison between data acquired with different hardware.

\subsubsection*{Environment-Change Generation Equipment}

To generate data with the defined parameters, we stabilize the following environment elements:

\begin{description}
	\item[Background] A solid color and static background ease the process of detecting facial features and extracting information from the surrounding environment. The background should ideally be black (or very dark) to prevent interference with the IR camera (black color has lowere reflectance compared to lighter colors)
	\item[Lighting] The room must be lit up by homogeneous light, and not produce shadows or glitters in the subject's face. By taking these measure, we ensure that the skin color will have no variations throughout the acquisition process.
\end{description}

\subsubsection{Acquisition Setup}

\begin{figure}
	\centerline{\includegraphics[width=0.7\linewidth]{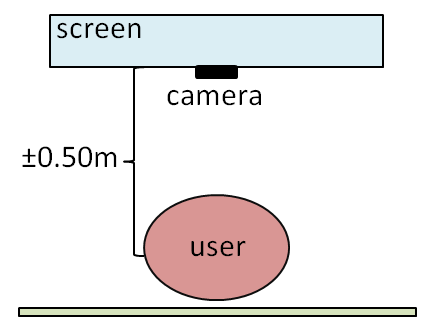}}
	\caption{Acquisition setup proposed for Protocol I.}
	\label{imgFdMieeSetup}
\end{figure}

The subject sits in front of the acquisition hardware. This hardware setup is composed of three cameras (LR, HR and IR). The subject's backdrop should be black with some space between them, to have the possibility of moving objects or subjects behind the main scene. This setup is exemplified in Figure \ref{imgFdMieeSetup}

\subsubsection{Protocol Guidelines}

To perform the acquisition, we suggested the presence of two members: one to perform the acquisitions (A) and the other to perform environment variations (B). The subject sits in front of the computer monitor and one of the team members aligns them with the cameras. During the entire acquisition procedure, the subject should remain as still as possible, to avoid producing changes during the various acquisition procedures.

Before starting the experiment, each subject has access to a printed copy of the protocol. This reduces the acquisition time, since the subject already knows what is going to take place during the experiment. Each performance parameter simulated and introduced in the scenario has its own guidelines:

\begin{description}
	\item[Control] Team member A takes a photo with the subject in the neutral face.
	\item[Lighting] Team member A takes 3 photos with different exposures (High, Medium, Low). This variable was only acquired in HR camera, because it is the only where it is possible to change the exposure level.
	\item[Background] Team prepare the background to the acquisition.
	\begin{enumerate}
		\item Team member A starts recording;
		\item Subject stay still during 5 seconds while team member B performs movement if necessary (only case of dynamic background);
		\item Team member A stops recording.
	\end{enumerate}
	\item[Multi-Subject] While subject is being record, team member B appear in the scene during 10 seconds.
	\item[Occlusions] For total occlusion, subject will start in the center of the scene and will slowly move to a point out of the scene. For partial occlusions, a photograph is taken with a plain color surface, like a piece of paper covering the following parts of the face:
	\begin{itemize}
		\item Top;
		\item Left;
		\item Bottom;
		\item Right.
	\end{itemize}
	\item[Head Rotation] For each head pose (Yaw, Pitch and Roll) subject performs the movement in both directions while being recorded through the complete movement.
	\item[Universal-Based Facial Expressions, plus Contempt] Subject repeat during 10 seconds the following emotion expressions, starting from the neutral pose to a full pose:
	\begin{itemize}
		\item Joy;
		\item Anger;
		\item Surprise;
		\item Fear;
		\item Disgust;
		\item Sadness;
		\item Contempt.
	\end{itemize}
	\item[Speech Facial Expressions] The subject reads a cartoon or text and is encouraged to express his feelings about it.
	
\end{description}

\subsubsection{Obtained Outputs}

This protocol generates the following output data:

\begin{itemize}
	\item HR and LR Photographies (.jpeg)
	\item LR camera videos - 15fps (.wmv)
	\item HR camera videos - 25fps (.mov)
	\item IR camera videos - 100fps (.avi)
\end{itemize}

The emotions generated through variation of facial parameters are expected to contain a mixture of macro and micro (i.e. subjects can be repressing and suppressing feelings) as well as false (i.e. subject is making an effort to express certain emotions) and subtle (i.e. when subject cannot generate a high intensity expression) plus speech-based expressions.

\subsubsection*{Data Organization and Nomenclature}

For standardization purposes and further analysis, a folder for each acquisition hardware was created. Inside these folders exist sub-folders for each of the tested performance parameters. The output files were placed in the respective folder with the following template naming convention: 

\begin{center}
	\textit{CaptureModeVolunteer0X\_SimulationName\_take0Y.format}
\end{center}

, where \textit{CaptureMode} is the type of hardware, \textit{X} is the number of the subject, \textit{SimulationName} is the name of the performance parameter acquired and respective information and \textit{Y} is the take's identification number.

\subsubsection{FDMiee Acquisition \& Protocol Validation}

Following the described protocol guidelines, we acquired data from eight volunteers with the following subject characteristics:
\begin{description}
	\item[Gender] Male/Female;
	\item[Glasses] With/Without;
	\item[Beard] With different formats/Without.
	\item[Age] 20-35 years
\end{description}

%In Figures \ref{imgControl}, \ref{imgExposure}, \ref{imgBackground}, \ref{imgPoses}, \ref{imgMacro}, \ref{imgSubtle}, \ref{imgMulti} and \ref{imgOcclusions}, it is possible to see sample results from each performance parameters with the different acquisition hardware.

\begin{figure}
	\centerline{\includegraphics[width=0.9\linewidth]{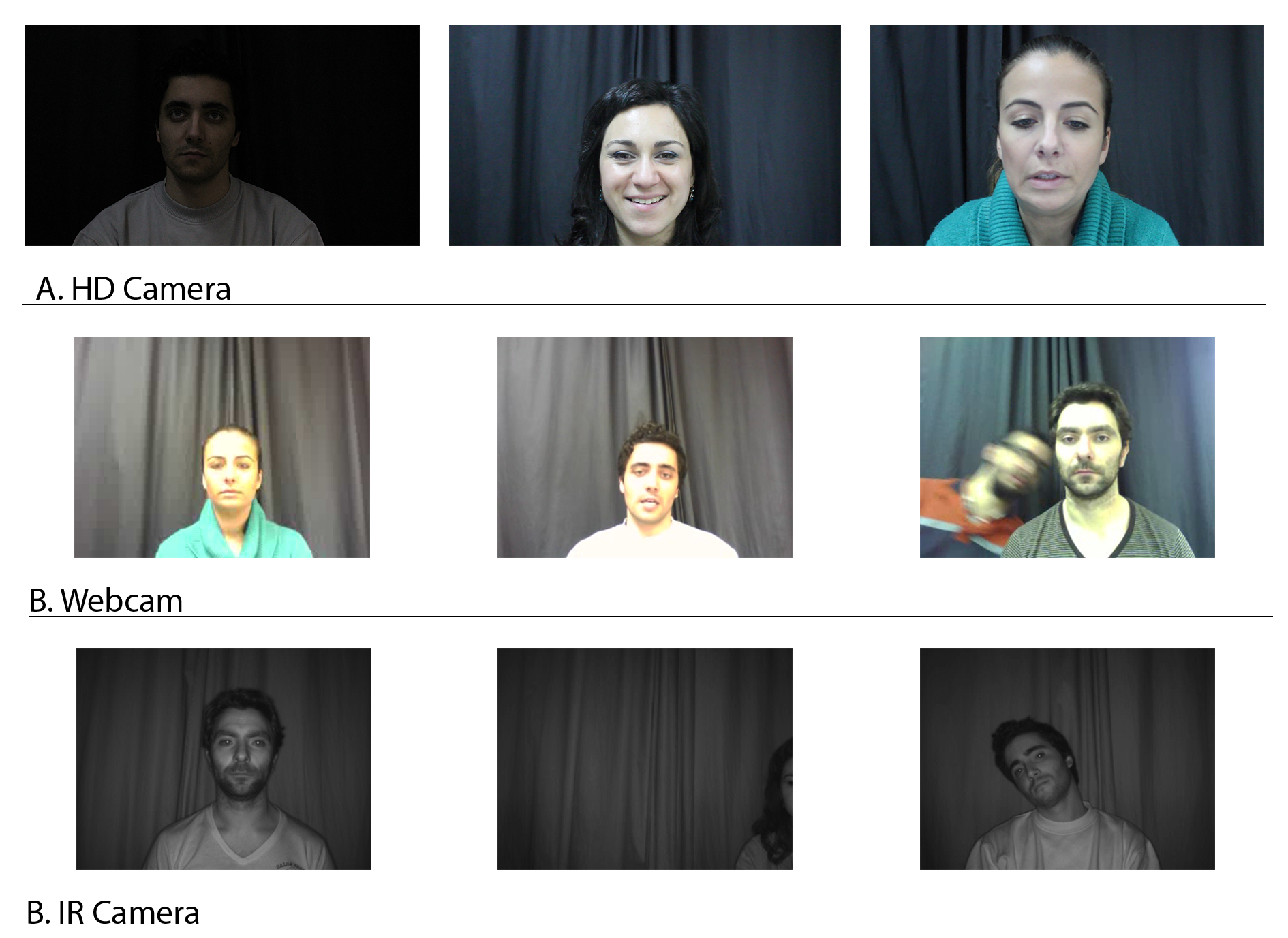}}
	\caption{FDMiee samples results for HD Camera (A), Webcam (B) and IR Camera (C)}
	\label{imgFDMieeResults}
\end{figure}

In Figure \ref{imgFDMieeResults} it is possible to see sample results from some of the performance parameters with the different acquisition hardware.

\subsection{Protocol II}

The definition and extraction of induced facial behaviors and speech features inherent to spontaneous expressions is still a challenge for CV systems. To develop and subsequently evaluate a CV algorithm that achieves this goal, we proposed these two protocols to acquire a database containing, simultaneously, spontaneous facial expressions and speech information inherent to induced emotions, such as Ekman's universal emotions \cite{ekman,martin2009philosophy}. Therefore, in this experiment we focus on the definition of guidelines to capture facial parameters changes in the performance variable \ref{tblDefinitionProtocol}.

\subsubsection{Requirements}

We define two types of requirements: emotion induction method and equipment requirements. Emotion induction method is used as basis to define the protocol guidelines inherent to facial parameters simulation.

\subsubsection*{The Emotion Induction Method}

The majority of spontaneous facial expressions are generated in real-life situations. To simulate these facial behaviors, we proposed a protocol where the system would ask for emotional acting in order to trigger facial responses from a subject. For this purpose, we combined a Mood Induction Techniques 1 (MIT 1) described by Hesse A.G. \textit{et al.} \cite{gerrards1994experimental} with mood induction sentences suggested by Pitas I. \textit{et al.} \cite{martin2006enterface}. As an application example, we could have a system that asks for certain user emotions expression through facial or speech features. The user must pronounce certain sentences with a particular tone and facial expression, matching the required emotional state. According to expression classification introduced in Section I, using this method we are able to induce macro, micro, false, masked, and subtle expressions. Macro expressions are implicit, since we ask for expression of the six of the Ekman universal emotions (i.e.  anger, fear, sadness, disgust, surprise and happiness). However, since we are in an induced emotions context, subjects can have difficulty engaging in the proposed situation and generating micro, false and masked expressions. Also subtle expressions are triggered because subjects' engaging intensity can be low in the induced sentence or context. As expected, the produced facial expressions depend of subjects' interpretation and how they emerge themselves in the simulated situation.

Our induction approach presents a novel view on emotion acting and their applications, though the domain still remains unexplored in state-of-the-art databases.

We used common persons as subjects, instead of actors, to maintain the natural-ness of real-life scenarios and also achieve a larger diversity of facial behaviors. Actors gain, over time, professional skills that common population cannot reproduce, thus they might introduce features that cannot match the real-world human performance. Some available databases that use MIT 1, try to induce emotions in the users, asking them to imagine themselves in certain general and predefined situations \cite{martin2006enterface,wilting2006real,velten1968laboratory}. We also avoided this approach, since suggesting certain situations will not guarantee certain emotion expressions as output by the subject. This is due to the fact that different individuals have different reactions as responses. Therefore, in our protocol, we asked the subjects to imagine and create for themselves some personal mental situation, while they enact the pre-defined sentence. This aims to ensure an engaging adaptation and natural response from the subject. As mentioned before, the chosen emotions were the six basic Ekman emotions \cite{ekman}, due to their scientific acceptance and applicability in real-world situations. The sentences are pronounced in the European Portuguese language to match the user’s mother tongue. This is another contribution of our work, since currently there is not a Multimodal European Portuguese database available. 

\subsubsection*{Equipment and Environment Requirements}

\begin{figure}
	\centerline{\includegraphics[width=0.5\linewidth]{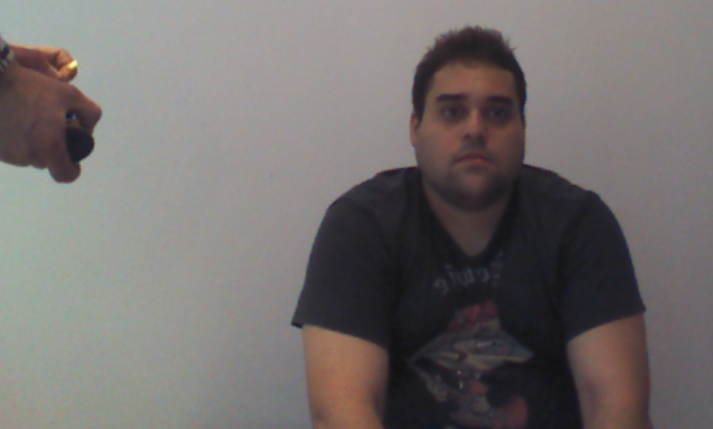}}
	\caption{Example of our video-audio synchronizer (left).}
	\label{imgFACIAsynch}
\end{figure}

The acquisition setup uses the Microsoft Kinect as acquisition hardware variable. Kinect records 3D data stream as well as speech information. The illumination is not controlled however, as acquisitions were executed during different day periods under uncontrolled lighting conditions. The background is static and white, and there is no sound isolation, since speech signal can be affected by external noise. Sentences are displayed on a screen positioned in front of the subject. To allow further synchronization or re-synchronization, a sound and light emitter is used in the beginning of each recording (see example of Figure \ref{imgFACIAsynch}). For this experiment in our protocol validation, we developed a software that allows simultaneous recording of color and depth video with speech from Microsoft Kinect, in .bin, video, and audio formats. This software includes the Facetracker's Microsoft SDK, and also saves the information retrieved from this algorithm.

\subsubsection{Acquisition Setup}

\begin{figure}
	\centerline{\includegraphics[width=0.75\linewidth]{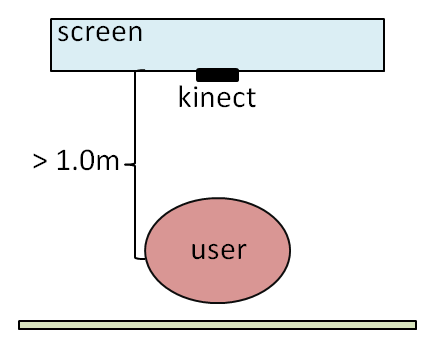}}
	\caption{Acquisition setup proposed by FACIA protocol}
	\label{imgFACIAsetup}
\end{figure}

The subject sits in front of the capture hardware. Distance between subject and Microsoft Kinect should be more than 1 meter to enable facial depth capture. A screen displays the sentence that is currently going to be "acted". The subject did not watch the recordings neither observe their own acting, to avoid auto-evaluation or influence their acting performance and expressivity. In FACIA protocol we propose the acquisition setup of Figure \ref{imgFACIAsetup}.

\subsubsection{Protocol Guidelines}

Each subject sits in front of the screen and acts out the two sentences per emotion \ref{tableFaciaSentences} while their voice and face expression are recorded. Per sentence we execute the procedure two times. This ensures the integrity of final results. We suggest a minimum of two members (A and B) in the acquisition team. Before starting the experiment, a protocol describing the experiment is given to the subject.

The experiment starts by a neutral sentence \cite{beaucousin2007fmri} (that can be used as baseline for further experiments). 

\begin{table}%
	\caption{FACIA emotion induction method: Sentences pronounced and acted by the subjects.\label{tableFaciaSentences}}{%
		\begin{tabular}{|l|l|}
			\hline
			\textbf{Emotion} & \textbf{Sentences} \\ \hline \hline
			Neutral & A jarra est\'a cheia com sumo de laranja  \\ \hline
			\multirow{4}{*}{Anger} 	& O qu\^e? N\~ao, n\~ao, n\~ao! Ouve, eu preciso deste \\
									& dinheiro! \\ 
									& Tu \'es pago para trabalhar, n\~ao é para beberes \\
									& caf\'e. \\\hline
			\multirow{2}{*}{Disgust}	& Ah, uma barata! \\ 
										& Ew, que nojo!\\\hline				
			\multirow{3}{*}{Fear}	& Oh meu deus, está algu\'em em minha casa! \\ 
									& N\~ao tenho nada para si, por favor, n\~ao me \\
									& magoe!\\\hline
			\multirow{2}{*}{Joy}	& Que bom, estou rico! \\ 
									& Ganhei! Que bom, estou t\~ao feliz!\\\hline    
			\multirow{2}{*}{Sadness}	& A minha vida nunca mais ser\'a a mesma. \\ 
										& Ele(a) era a minha vida.\\\hline       
			\multirow{2}{*}{Surprise}	& E tu nunca me tinhas contado isso?! \\ 
										& Eu n\~ao estava nada \`a espera!.\\\hline      
		\end{tabular}
	}
	%		\begin{tabnote}%
	%			\Note{Source:}{This is a table
	%				sourcenote. This is a table sourcenote. This is a table
	%				sourcenote.}
	%			\vskip2pt
	%			\Note{Note:}{This is a table footnote.}
	%			\tabnoteentry{$^a$}{This is a table footnote. This is a
	%				table footnote. This is a table footnote.}
	%		\end{tabnote}%
\end{table}%

Therefore, to each sentence of Table \ref{tableFaciaSentences} the following pipeline is repeated two times:

\begin{quote}
	\begin{enumerate}
		\item Acquisition team member A says \textit{1,2,3… I will record!}.
		\item Acquisition team member B uses the light/sound synchronizer.
		\item Subject performs the emotion acting.
		\item Acquisition team member A stops the recording.
	\end{enumerate}
\end{quote}

\subsubsection{Obtained Outputs}

Using our acquisition protocol, we obtained the following data per sentence enacted:

\begin{itemize}
	\item Video Color – Resolution - 30fps (.bin);
	\item Depth image – Resolution - 30fps (.bin);
	\item Audio – pcm format – 16000 Hz (.bin).
	\item Facetracker SDK and Action Units detected (.bin).
	\item Audio file (.wave).
	\item Color Video file (.avi).
\end{itemize}

As explained, regarding facial behaviors we are able to generate data containing macro, micro, false, masked and subtle expressions.

\subsubsection*{Data organization and Nomenclature}

Similarly to procedure adopted in FdMiee protocol, we predefine how data acquire is going to be organized. To each subject is created a folder called “Volunteer0X”, where X is the number associated to the subject. Inside each subject folder are created eight additional folders: one per emotional sentence. Inside of each emotion folder we will have two folders numbered with corresponding sentence, where we will place three data types obtained.
Regarding file names, we will use the following template:
\begin{center}
	\textit{ Volunteer0XEmotionSentence0YTake0Z.format}
\end{center}

Where X is the subject number, Y the sentence number and Z the take number.

\subsubsection{FACIA acquisition \& Protocol validation}

\begin{figure}
	\centerline{\includegraphics[width=0.9\linewidth]{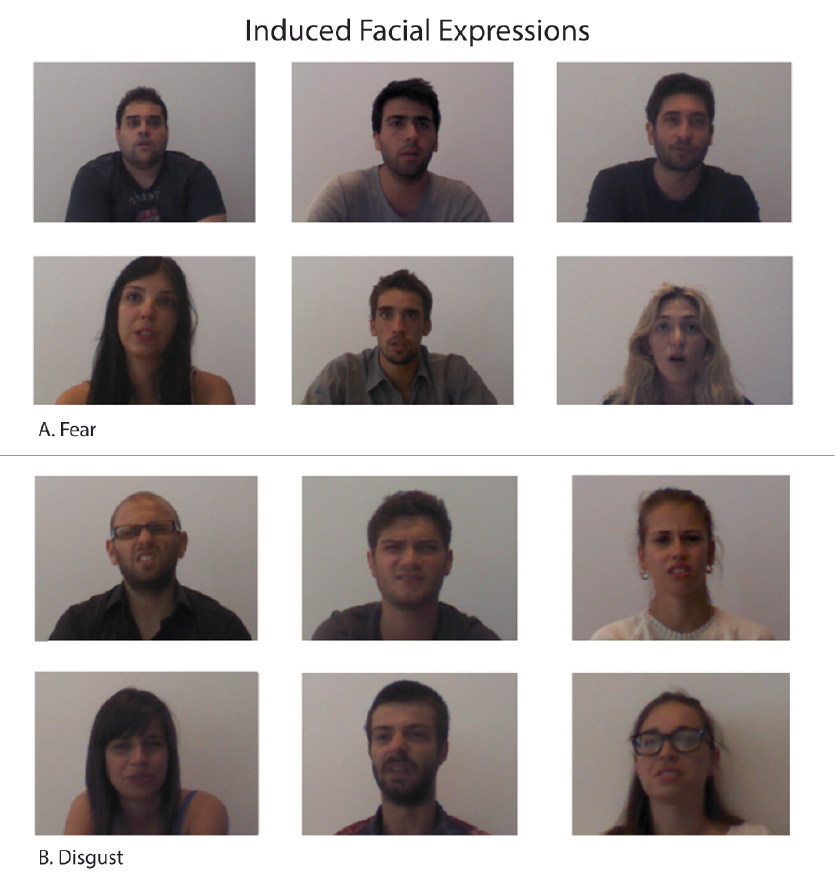}}
	\caption{Sample of results obtained for fear (A) and (B) disgust emotion acting}
	\label{Figures_v04_F_FACIA_sample_results}
\end{figure}

To validate Protocol II, we follow it for eighteen subjects, in a total of 130 files per subject (total of 504 acquisitions). As subject characteristics variable we have seven female and eleven male; ages are in a range of 20-35 years old and they were all caucasian. As already explain, we require depth information so a Microsoft Kinect was used as acquisition hardware. As sample of results acquired during validation we can observed the Figure \ref{Figures_v04_F_FACIA_sample_results}.

\section{Discussion and Conclusions}

In this paper, we presented a methodology to facilitate the development of two facial data acquisition protocols. Following this methodology, we presented the protocols for simulation and capturing of real-life scenarios and facial behaviors. To validate the protocols, two sample databases were created: FdMiee and FACIA. They contain comprehensive information on facial variations inherent to both spontaneous and non-spontaneous facial expressions under a wide range of realistic and uncontrolled situations. Generated databases can be used in a variety of applications, such as CV systems evaluation, testing, and training \cite{baggio2012mastering}. They also serve as proof-of-concept. Adopting our methodology and following our protocols reduces the time required for customized database acquisition.

Throughout the protocol creation process, we characterized two groups of variables: flexible variables (subjects' characteristics and capture hardware) and fixed performance variables (external and facial parameters). The first protocol focuses on external parameters' simulation as variation of the fixed performance variable. As an extension, the second protocol provides guidelines to induce and capture real-life facial behaviors as fixed performance variables.

Protocol I allows the acquisition of a facial database containing a large number of fixed parameters' variations (external and facial): lightning, background, multi-subject, occlusions, head rotation, universal-based, and speech facial expressions (Table \ref{tblDefinitionProtocol}). Lighting variations introduce changes in facial features (e.g. contrast and brightness) \cite{KCLee05}. These variations enable us to test how CV systems react to and detect, and how tracking is affected. Static and dynamic variations in the background usually interfere with CV systems' performance while detecting and tracking faces \cite{hsu2002face}. Therefore, in this protocol, we simulate different background contexts, as well as introduce static and dynamic features in the environment. Similar to background variables, we simulate multi-subject environments, since this situation usually interferes with, and at times, disables CV systems' feature detection \cite{baggio2012mastering}. Occlusions generated by glasses or hardware are also common in real-life scenarios, influencing face recognition and emotion classification accuracy \cite{Cotter2010,Buciu2005,Bourel2001}. The increase of Head-Mounted-Displays usage in Virtual Reality applications makes it crucial to test systems invariance while using these variables. Regarding facial behaviors, we reproduced and captured two kinds of facial behaviors - universal-based and speech-based facial expressions. Universal-based Facial Expressions are related to pure emotions \cite{ekman}. They provide data for emotion recognition systems and enable the testing of systems invariance while subjects' faces change expressions. Speech Facial Expressions, on the other hand, are inherent to all types of expressions \cite{Koolagudi2012sim} (as showed in the image \ref{imgFacialDatabasesWorld}) and enable the measuring of systems accuracy and precision. To validate Protocol I, we performed an acquisition on eight subjects with different subject characteristics, leading to the creation of FdMiee database. FdMiee contains facial behaviors under different environment contexts. Hence, this protocol enables the generation of databases that are useful for a wide range of CV systems performance tests.

Protocol II extends the first protocol regarding facial behaviors and performance variables, by introducing induced facial features. To achieve this, we proposed an emotion induction method, where facial expressions were induced through emotional acting. Analysing FACIA generated in the validation process, we verified that facial behaviors inherent to certain emotional acting are indeed different among individuals; i.e. subjects performed different acts to realize identical emotional states. Analysing subjects' facial behaviors, we were able to simulate all types of expressions according to subjects interpretations and engaging in induction sentences. Hence, this protocol provides a large and heterogeneous set of facial behaviors, useful for determining the accuracy of tracking and recognition systems. This was intuitively expected, since expressions inherent to emotional states share some action units \cite{ekman}. This mixing of expressions can compromise database usage to train a machine learning classifier in pure expressions recognition, increasing classification error. Microsoft Kinect was chosen as the acquisition hardware variable, so that we could record three kinds of data: color, depth (3D facial information) and speech. Introducing depth in the stored data provides valuable information \cite{Zhang2014692}. However, recent studies point out that acquisition rate of Kinect is not sufficient for micro and subtle expressions capturing \cite{ekman,martin2009philosophy}. This argument explains the poor component of micro and subtle expressions present in FACIA. However, in our methodology we classify this variable as flexible, to ensure that protocol guidelines can be used with other acquisition hardware, i.e. guidelines can be applied with high frame rate cameras and improve the capture of these facial behaviors. The speech recording also allows the Portuguese emotional data collection, opening novel research lines in emotion classification and recognition present in the European Portuguese language speech.

In conclusion, our proposed methodology facilitate the generation of facial data acquisition protocols. This methodology provides a tool for researchers to develop their own facial databases. It also enable performance tests, validation and training processes in CV systems in a wide range of life-like scenarios and facial behaviors, being adaptable to different subject characteristics and acquisition hardware. 

\section{Future Work}

Our further work will focus on the following key tasks: First, we aim to enlarge our proof-of-concept sample databases and, subsequently, perform a statistical validation of the two protocols presented in this paper. Enlarging the databases will provide sufficient data for statistical validation, using various CV systems. The statistical validation will also provide more measurable information regarding data significance and impact. Second, we aim to devise more parameters for methodology variables to refine the validation process. Third, we aim to introduce a more heterogeneous subject samples, with a wider age range (thus greater presence of wrinkles and facial pigments), skin colors, and make-up. Fourth, we intend to carry out tests with more sophisticated acquisition hardware, such as high-speed cameras. And finally, to increase our work applicability, we intend to extend the fixed variable of performance parameters, providing more guidelines to generate novel situations.



%% References


\bibliographystyle{model3-num-names}
\bibliography{elsarticle-template-3-num-cag}

%% not want to use model3-num-names.bst.





% \end{thebibliography}

\end{document}